\documentclass[journal]{IEEEtran}
\usepackage{amsmath,amssymb,amsfonts}
\usepackage{array}
\usepackage{tabularx}
\usepackage{graphicx}
\usepackage{booktabs}
\usepackage{multirow}
\usepackage{textcomp}
\usepackage{siunitx}
\usepackage{float}
\usepackage{stfloats}
\usepackage{url}
\usepackage[numbers,sort]{natbib}
\usepackage[hidelinks]{hyperref}

\begin{document}

\title{TRNet: Learning with Topographic Priors for VHR Paddy Rice Mapping}

\author{Kaiwen Xiao, Chunlong Fu, Liping Zheng, Yanfeng Su, and Yuanhao Xi%
\thanks{We acknowledge support by the Open Access Publication Funds of the
University of G\"ottingen. This work was supported by the Youth Science Fund
Project of Sichuan University Jinjiang College under Grant QNJJ-2025-A05.
(Corresponding author: Yuanhao Xi.)}%
\thanks{Kaiwen Xiao, Chunlong Fu, Liping Zheng, and Yanfeng Su are with the
School of Computer Science, Sichuan University Jinjiang College,
Meishan 620860, China
(e-mail: xiaokaiwen@scujj.edu.cn; fuchunlong@scujj.edu.cn;
zhengliping@scujj.edu.cn; su2997159319@outlook.com).}%
\thanks{Yuanhao Xi is with the Institute of Computer Science,
University of G\"ottingen, 37077 G\"ottingen, Germany
(e-mail: yuanhao.xi@stud.uni-goettingen.de).}}

\maketitle

\begin{abstract}
Mapping paddy rice from very high resolution (VHR) imagery in mountainous and hilly regions remains challenging because terrain variations alter optical appearance and increase confusion with visually similar vegetation. To address this issue, we propose TRNet for multimodal paddy rice segmentation using 0.5 m GaoJing 1 red green blue (RGB) imagery, a 5 m TanDEM X digital elevation model (DEM), and derived slope information. TRNet employs separate visual and terrain encoders to preserve modality specific representations. At an early encoder stage, the proposed Topographic Energy Spectral Rectification (TESR) performs terrain conditioned low frequency modulation and asymmetric high frequency regulation to suppress steep slope clutter while selectively enhancing rice related cues on compatible low slope regions. The Topography Guided Paddy Structure Decoder (TPSD) further integrates semantic, rice background boundary, and interior cues with coarse topographic context to refine structural predictions. Experiments are conducted on an Area A internal test set and a geographically held out Area B with steeper terrain and lower rice prevalence. TRNet achieves Rice IoU scores of 85.10\% and 80.68\% on Areas A and B, outperforming the original Dual Encoder U Net by 9.15 and 18.83 percentage points, respectively. Without any adaptation, evaluation on matched August 2024 imagery retains Rice IoU scores of 82.04\% and 76.12\%. Extensive ablation, slope stratified, and cross year seasonal analyses demonstrate that the improvements arise from effective frequency rectification and structure learning, which reduce steep terrain false positives and low slope rice omissions. These results demonstrate that coarse topography can serve as a stable contextual prior for robust VHR paddy rice mapping.
\end{abstract}

\begin{IEEEkeywords}
Paddy rice mapping, very-high-resolution imagery, topography-guided learning, multimodal semantic segmentation.
\end{IEEEkeywords}

\section{Introduction}

Accurate paddy rice mapping is essential for agricultural management, irrigation planning, crop statistics, and regional monitoring~\citep{wei2022rice,yu2023ricemapengine,zhang2025paddyclassification,xu2025countylevelrice}. Remote sensing has become a primary tool for large scale rice monitoring because it enables efficient and repeatable observation over broad geographic regions~\citep{jeongDevelopmentVariableThreshold2012,gaoFARMFullyAutomated2023,ni2021enhanced,pan2025highly}. With the increasing availability of very high resolution (VHR) imagery, paddy rice mapping is further moving from regional identification toward parcel level delineation, where field boundaries, parcel geometry, and within field texture become directly observable~\citep{zhangParcellevelMappingCrops2020}. Such fine grained spatial information can be effectively modeled by modern encoder decoder segmentation networks~\citep{ronnebergerUNetConvolutionalNetworks2015,diakogiannis2020resunet}. However, the advantages of VHR imagery also expose stronger local appearance variations, making reliable rice mapping considerably more difficult in mountainous and hilly regions.

The difficulty arises from two coupled factors. First, topographic variation changes illumination and land cover appearance, while mountain shadows introduce strong radiometric discontinuities~\citep{chen2023topographic,song2026gan}. Under VHR observation, these effects appear as salient textures and contrasts that may resemble vegetation or field structures~\citep{cui2023siamc}. 
As a result, segmentation models can respond strongly to steep non rice terrain, making terrain induced visual detail difficult to distinguish from genuine rice patterns~\citep{zhao2026high}. 
Second, rice fields in mountainous areas are often small, fragmented, and irregularly distributed~\citep{liu2024regional,wang2024small}. Their narrow interiors and complex boundaries are easily broken during prediction, leading to incomplete contours and inconsistent region interiors even when multiscale or edge aware architectures are adopted~\citep{waldner2020deep}. Therefore, robust VHR paddy rice mapping requires not only better discrimination between terrain induced appearance and true rice structure, but also preservation of fragmented field geometry.

Existing studies mainly address the first issue through multisource fusion or frequency domain modeling. Multisource methods introduce auxiliary modalities to complement RGB observations. The Gather to Guide Network aggregates multimodal features to guide RGB representations~\citep{zheng2021gather}, while CMX rectifies modality specific features before cross modal fusion~\citep{zhangCMXCrossModalFusion2023}. For optical and topographic data, DMTFNet separately encodes Sentinel 2 imagery and terrain features and combines them through multiscale Transformer decoding~\citep{gao2026dual}, whereas FCA DeepLab integrates optical and DEM streams using spatial and channel attention~\citep{tuo2026landslide}. These methods demonstrate the value of complementary modalities, but directly fusing coarse DEM or slope with VHR imagery introduces a fundamental spatial scale mismatch. Upsampled topographic data can provide terrain context, yet they cannot recover field scale spatial detail that is absent from the original elevation source. More broadly, recent conditional visual modeling has shown that auxiliary priors can effectively regulate visual representations when their roles are explicitly defined rather than simply concatenated~\citep{shen2024advancing,shen2024imagpose,shen2025imagdressing,shenlong}. This observation motivates us to treat topography as contextual guidance rather than as another fine resolution visual source.

Frequency domain modeling provides a perspective by separating feature responses according to spatial frequency. SFFNet combines wavelet derived frequency features with spatial representations~\citep{yang2024sffnet}, while hierarchical wavelet enhancement strengthens frequency components at network stages~\citep{li2023waveletEnhancement}. Dual domain methods further couple wavelet features with spatial constraints~\citep{wei2024dualDomain}, and learned frequency fusion extends this interaction to heterogeneous remote sensing modalities~\citep{chen2025learning}. Although these methods explicitly regulate scale dependent information, frequency decomposition alone cannot determine whether high frequency responses correspond to genuine field structure or terrain induced clutter. In mountainous scenes, this distinction is crucial because steep slopes and shadows can generate strong high frequency responses that resemble crop boundaries. Therefore, frequency regulation requires an explicit topographic criterion that determines when visual detail should be suppressed or preserved.

The second issue, incomplete boundaries and inconsistent interiors, has mainly been addressed through structure aware supervision. E2EVAP combines semantic contour interaction with topological constraints for parcel boundary extraction~\citep{pan2023e2evap}. BsiNet jointly predicts parcel masks, boundaries, and distance maps~\citep{long2022delineation}, while BSNet fuses semantic and boundary representations for farmland parcel mapping~\citep{shunying2023bsnet}. Semantic edge aware multitask learning further improves region and edge representations through joint supervision~\citep{li2023using}, and boundary specific objectives explicitly penalize contour errors~\citep{kervadec2021boundary,wang2022activeBoundary}. These approaches improve geometric delineation, but structural supervision alone cannot resolve terrain induced ambiguity. A visually plausible boundary may still correspond to steep non rice terrain, while true rice interiors may be suppressed by local appearance changes. Coarse topography provides complementary evidence of rice terrain compatibility, yet how to couple such context with boundary and interior structure for joint refinement remains insufficiently explored.

To address these limitations, we propose the Topography Regulated Network (TRNet), an asymmetric multimodal framework that treats VHR imagery as the primary source of semantic and spatial evidence while using coarse topography to regulate visual interpretation. TRNet contains two key components. Topographic Energy Spectral Rectification (TESR) introduces terrain conditioned frequency regulation at an early encoder stage, where low frequency semantics are modulated by terrain context and high frequency responses are asymmetrically regulated to suppress steep slope clutter while preserving compatible rice cues. Topography guided Paddy Structure Decoder (TPSD) further couples semantic, boundary, and interior information with coarse topographic context to refine fragmented rice regions without treating terrain as fine boundary evidence. Through this design, TRNet explicitly addresses both terrain induced visual confusion and structural fragmentation within a unified framework.
The main contributions are summarized as follows:
\begin{itemize}
    \item We propose an asymmetric terrain as context framework for VHR paddy rice mapping, where RGB imagery provides the primary semantic and spatial evidence, while DEM and slope regulate visual interpretation through modality specific representations.

    \item We introduce TESR and TPSD to address two complementary challenges. TESR performs terrain conditioned low frequency modulation and asymmetric high frequency regulation to reduce terrain induced false responses, while TPSD integrates boundary, interior, and topographic cues to refine fragmented rice structures. Both modules require only binary rice masks for supervision.

    \item Extensive experiments on the Area A internal test set and geographically held out Area B demonstrate Rice IoU values of 85.10\% and 80.68\%, exceeding the original Dual Encoder U Net by 9.15 and 18.83 percentage points, respectively. Without adaptation, TRNet further retains 82.04\% and 76.12\% Rice IoU on matched 2024 imagery. Component, frequency, structure, terrain stratified, and cross year analyses consistently show reduced steep terrain false positives and fewer low slope rice omissions.
\end{itemize}

\begin{figure*}[t]
\centering
\includegraphics[
width=0.96\textwidth,
height=0.48\textheight,
keepaspectratio
]{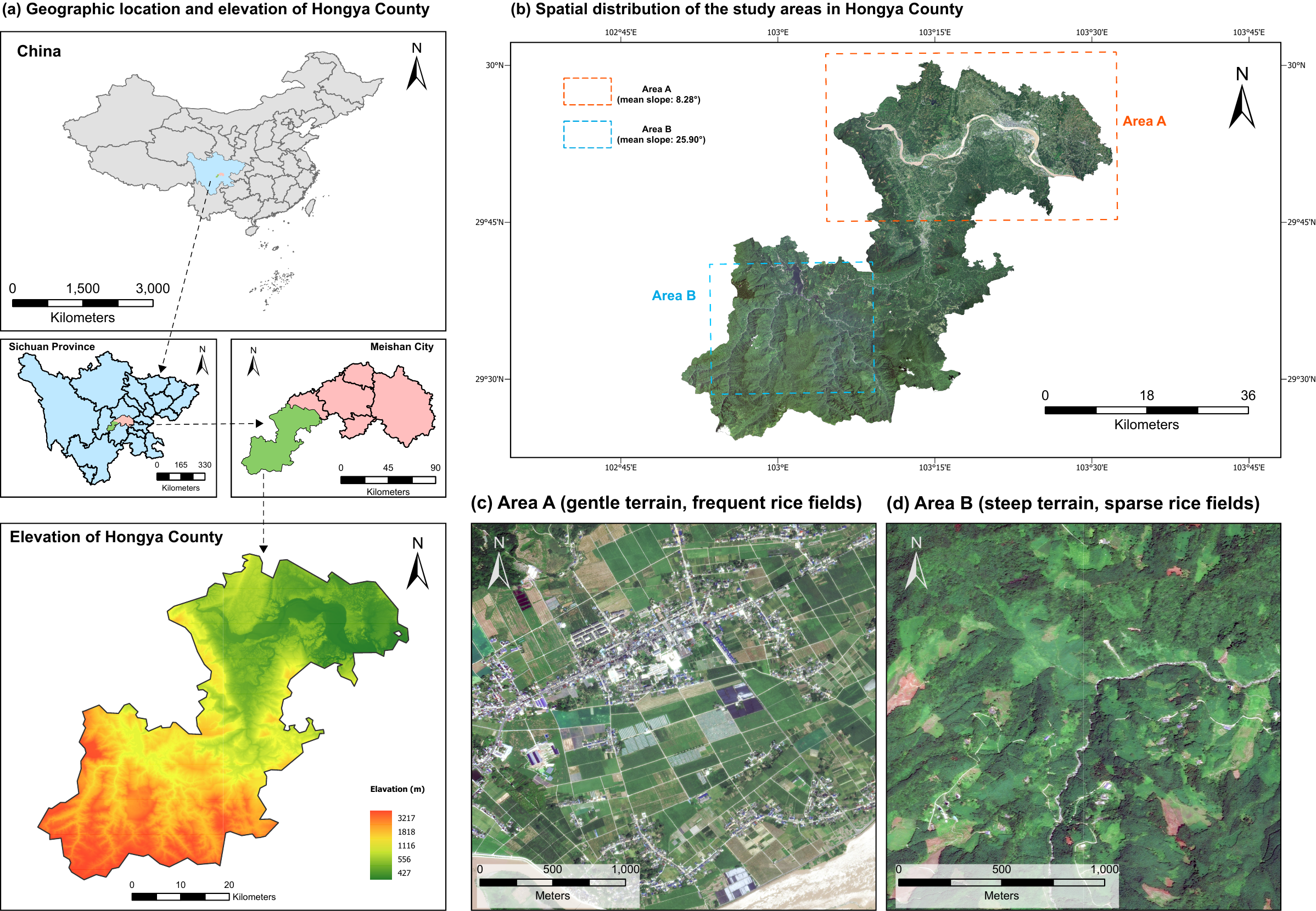}
\caption{Study areas in Hongya County, China: (a) county location and elevation, (b) Areas A and B, and representative 2023 GaoJing-1 imagery of (c) Area A and (d) Area B.}
\label{fig:study_area}
\end{figure*}

\section{Study areas and datasets}

\label{sec:study_areas_datasets}

\subsection{Study Areas}
\label{subsec:study_areas}

The study was conducted in Hongya County, Meishan City, Sichuan Province, China, a representative mountainous and hilly region characterized by heterogeneous terrain and fragmented agricultural landscapes. To evaluate both within area performance and geographic generalization, we selected two spatially separated subareas, denoted as Area A and Area B, as shown in Fig.~\ref{fig:study_area}. Area A was used for model development, including training, validation, and internal testing, whereas Area B was completely excluded from model development and reserved for held out cross area evaluation. The two areas exhibit substantially different topographic and agricultural characteristics. Area A contains extensive agricultural land and relatively frequent rice fields, with a mean slope of $8.28^{\circ}$. In contrast, Area B is considerably steeper and contains fewer rice fields, with a mean slope of $25.90^{\circ}$. Consequently, evaluation on Area B reflects a combined shift in terrain, land cover composition, and rice prevalence rather than an isolated change in topography. This design provides a more challenging setting for assessing the robustness of paddy rice mapping under realistic geographic variation.

\subsection{Data Sources and Preprocessing}
\label{subsec:data_sources}

The primary optical data consist of orthorectified and pan sharpened $0.5$ m GaoJing 1 (SuperView 1) RGB imagery acquired in July 2023. To evaluate cross year seasonal generalization, we additionally collected GaoJing 1 imagery over the same spatial footprints in August 2024 within the corresponding rice phenological period. Using the same sensor, nominal resolution, preprocessing procedure, and spatial support reduces confounding effects caused by cross sensor and geographic differences. Topographic information was obtained from the commercial $5$ m TanDEM X digital elevation model acquired in July 2023, following the processing protocol of \citet{rossiTanDEMXCalibratedRaw2012}. Slope in degrees was derived from the DEM using ArcGIS Pro. All data were projected to WGS 84/UTM Zone 48N (EPSG:32648). DEM and slope were bilinearly resampled to $0.5$ m and co registered with the RGB imagery only for grid alignment, without introducing spatial detail beyond the original terrain resolution. Such explicit alignment of heterogeneous conditions is also consistent with recent controllable visual modeling studies that emphasize semantically and spatially coherent conditioning~\citep{shen2025imagedit,shen2025imagharmony,shen2026imaggarment}. The resulting five channel input was clipped for outliers and standardized using fixed training set statistics. Degree valued slope was retained separately for TESR, TPSD, and TCE in Section~\ref{sec:methods} to ensure that physically defined slope thresholds were applied to the original measurements.

\subsection{Dataset Construction and Split}
\label{subsec:dataset_split}

\begin{figure}[t]

\centering

\includegraphics[width=\columnwidth]{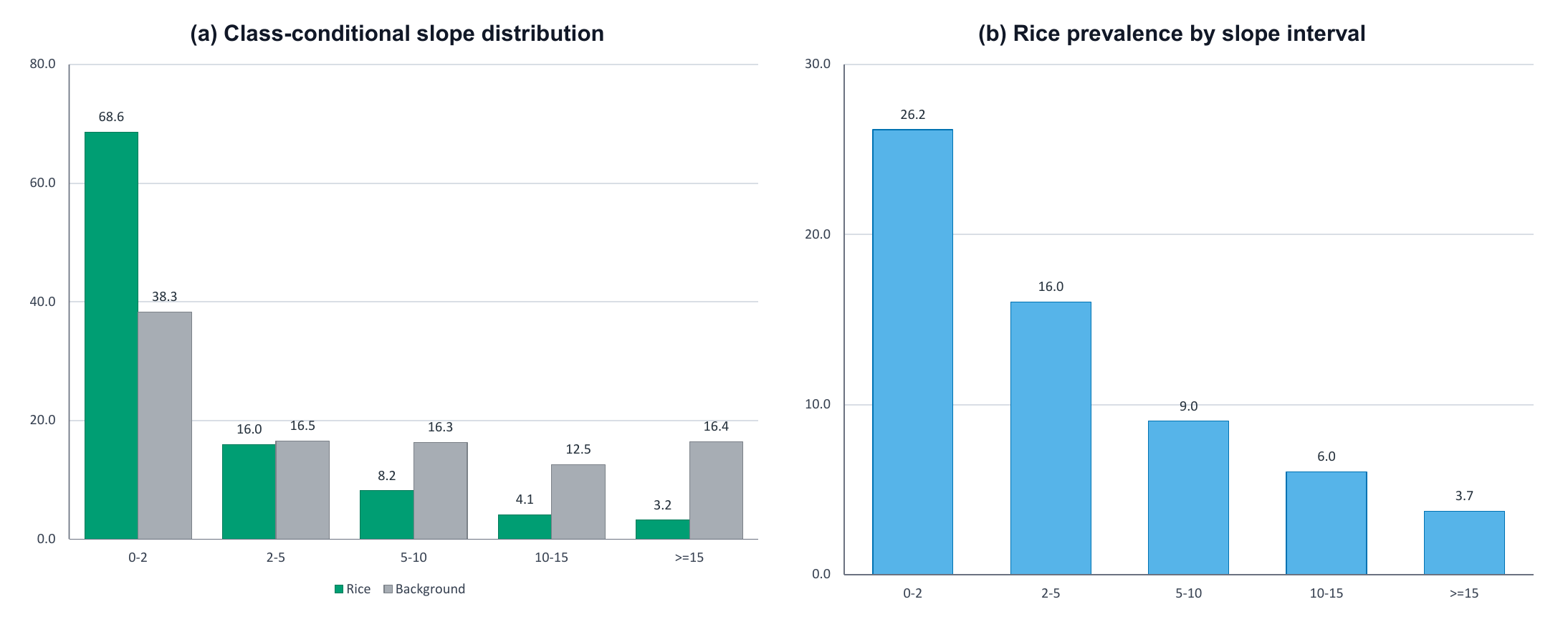}

\caption{Empirical relationship between slope and paddy occurrence in the Area A training split. (a) Class-conditional slope distributions, normalized separately for rice and background. (b) Fraction of rice pixels within each slope interval. Values are percentages over valid pixels.}

\label{fig:slope_distribution}

\end{figure}

Thirty annotators from an agricultural research institute manually interpreted the $0.5$ m RGB imagery without access to terrain information. The resulting annotations and co registered multimodal data were divided into 1,762 non overlapping patches of $512 \times 512$ pixels, each containing a five channel GeoTIFF and a binary rice mask. Area A contains 1,462 patches, which were divided into 1,169 training, 146 validation, and 147 internal test samples. Rice accounts for 16.51\%, 15.61\%, and 17.25\% of valid pixels in these three subsets, respectively. The split was fixed throughout all experiments. The remaining 300 patches belong to geographically separated Area B and were used exclusively for cross area testing, with rice occupying 4.80\% of valid pixels. For the August 2024 evaluation, the same 147 Area A and 300 Area B footprints were retained, while their masks were independently reinterpreted from the 2024 imagery to account for changes in planting status, fallow fields, and visible field boundaries. Rice occupies 16.38\% and 4.52\% of valid pixels in the two 2024 test sets. Neither the 2024 imagery nor its annotations were used for training, validation, normalization, checkpoint selection, or threshold adjustment. As shown in Fig.~\ref{fig:slope_distribution}, 68.6\% of rice pixels occur below $2^{\circ}$, whereas only 3.2\% occur at slopes of at least $15^{\circ}$. This strong but nonexclusive relationship motivates terrain aware regulation while avoiding the use of slope as a hard decision rule.

\section{Methods}
\label{sec:methods}

\subsection{Overall architecture of TRNet}
\label{subsec:overall_architecture}

\begin{figure*}[t]
\centering
\includegraphics[width=\textwidth]{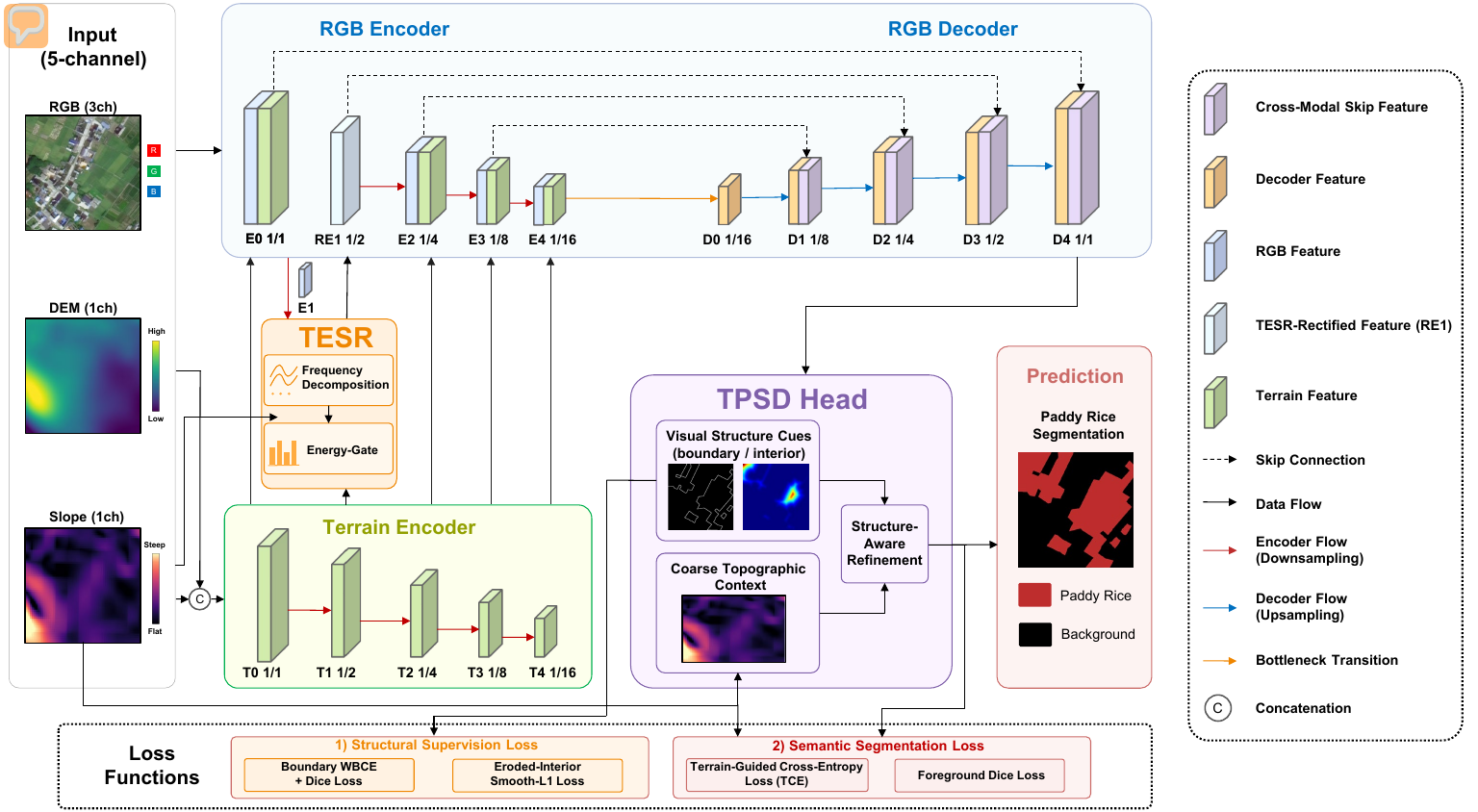}
\caption{TRNet architecture. RGB and DEM--slope streams feed five aligned encoder stages with 64, 128, 256, 512, and 1024 channels; degree-valued slope supports TESR, TPSD, and the Terrain-Guided Cross-Entropy Loss (TCE). Terrain-to-RGB fusion operates at $E_0$ and $E_2$--$E_4$. At $E_1$, TESR uses $T_1$ and degree-valued slope to produce $E_1'$ (shown as $\mathrm{RE}_1$) for $E_2$ and the decoder skip. The single RGB decoder combines the fused $E_4$ bottleneck with skips from $E_3$, $E_2$, $E_1'$, and $E_0$. TPSD refines the prediction using boundary, interior, and topographic cues. Training combines boundary WBCE + Dice, eroded-interior Smooth L1, TCE, and foreground Dice losses.}
\label{fig:framework}
\end{figure*}

TRNet is inspired by the dual-encoder paradigm of \citet{luDualEncoderUNet2023}. Both architectures use modality-specific encoders and asymmetric cross-modal interaction within a U-Net framework, but they assign different roles to the two modalities. In the original network, optical features enter the DEM companion encoder and the resulting companion features feed an attention-gated U-Net decoder. In TRNet, terrain features instead regulate the RGB pathway, allowing topography to provide contextual guidance while RGB remains the primary source of semantic and spatial evidence (Fig.~\ref{fig:framework}).

The RGB input has three channels, and the terrain stream contains standardized DEM and slope. Degree-valued slope is retained separately for TESR, TPSD, and TCE. The two encoders have five aligned stages with 64, 128, 256, 512, and 1024 channels. At $E_0$ and $E_2$--$E_4$, same-resolution RGB and terrain features are concatenated and projected back to the stage width. At $E_1$, TESR replaces this standard interaction and sends its rectified RGB feature to the next encoder stage and the corresponding decoder skip. The terrain encoder continues independently, while only the RGB pathway is decoded.

The four decoder stages start from the fused $E_4$ bottleneck, successively upsample by a factor of two, and combine skips from $E_3$, $E_2$, $E_1'$, and $E_0$. The resulting full-resolution 64-channel feature enters TPSD, which returns two-class segmentation logits together with one-channel boundary and interior-depth logits. The auxiliary logits are used only during training.

For controlled comparisons, \emph{TRNet-Base} uses the same asymmetric backbone and decoder but removes both proposed modules. It restores concatenation--projection at $E_1$ and replaces TPSD with a conventional $1\times1$ semantic classifier. TRNet-Base is therefore the architectural reference for module ablations, whereas the original Dual-Encoder U-Net is retained as a prior-art comparison.

\subsection{Topographic Energy-Spectral Rectification}
\label{subsec:tesr}

\begin{figure*}[t]
\centering
\includegraphics[width=\textwidth]{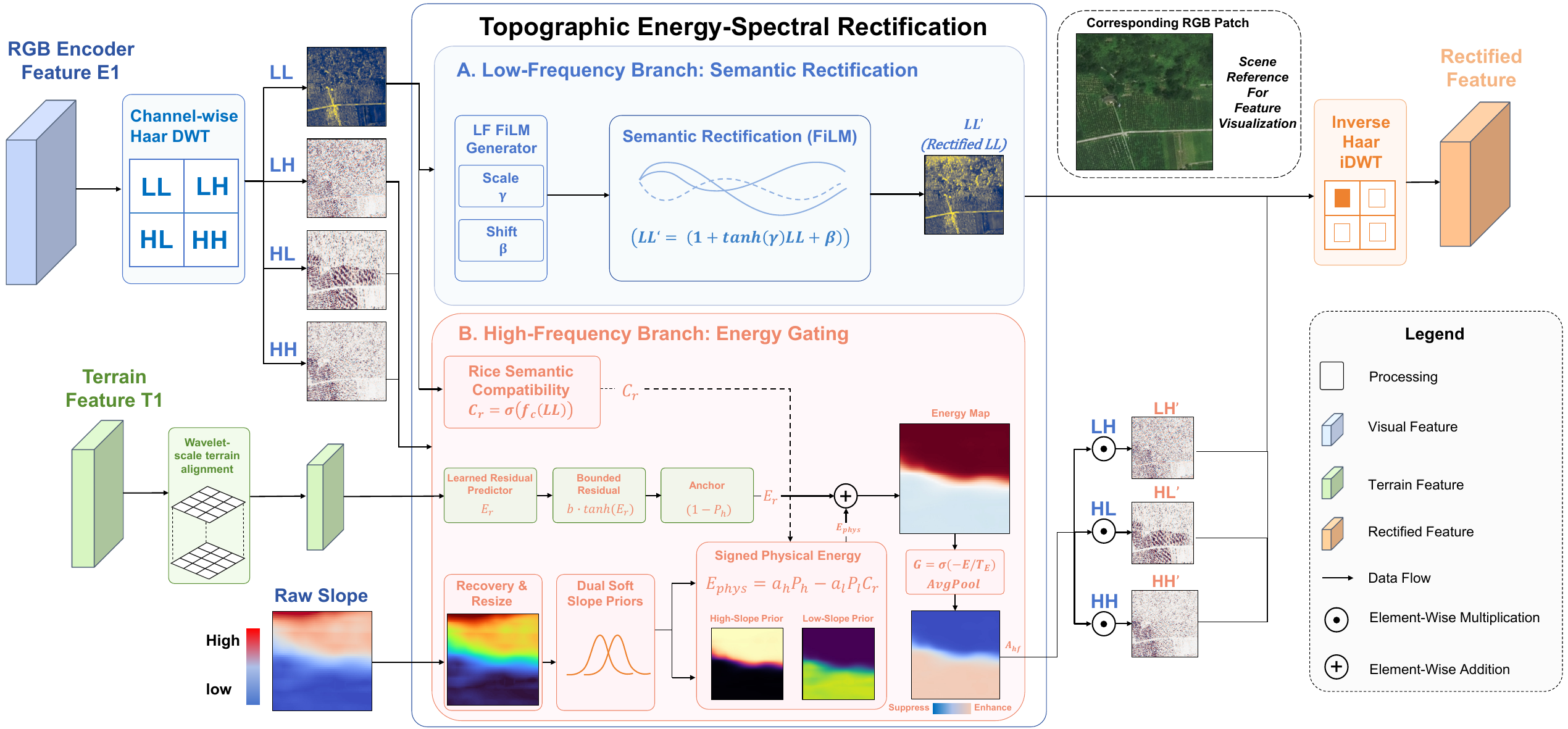}
\caption{Topographic Energy-Spectral Rectification (TESR) at the $E_1/T_1$ interaction point. Both feature inputs have shape $128\times h\times w$. Fixed Haar DWT decomposes $E_1$ into $LL$, $LH$, $HL$, and $HH$, each of shape $128\times h/2\times w/2$. A stride-2 terrain projection maps $T_1$ to $T_f$. The low-frequency path predicts $\gamma$ and $\beta$ for $LL'=(1+\tanh\gamma)\odot LL+\beta$. In the high-frequency path, high- and low-slope priors $P_h$ and $P_l$ are combined with visual compatibility $C_r$ and a high-slope-attenuated learned residual to form the terrain-guided energy $E$. The monotonically decreasing gate $G=\sigma(-E/T_E)$ is smoothed by size-preserving $3\times3$ average pooling to obtain $\overline{G}$, from which the centered gain $A_{\mathrm{hf}}=1+\alpha(2\overline{G}-1)$ is shared by $LH$, $HL$, and $HH$. Fixed Haar iDWT reconstructs the sole output $E_1'$, which proceeds directly to the next RGB encoder stage and the corresponding decoder skip.}
\label{fig:tesr}
\end{figure*}

TESR addresses two asymmetric terrain-related risks: texture-induced false responses on steep slopes and weak rice responses on flatter terrain. It suppresses terrain-inconsistent high-frequency responses at high slopes while conditionally enhancing compatible low-slope rice cues (Fig.~\ref{fig:tesr}). Existing remote-sensing segmentation methods use wavelets to combine spatial and frequency representations or enhance high-frequency subbands \citep{yang2024sffnet,li2023waveletEnhancement}, while cross-modal rectification methods use one modality to calibrate another \citep{zhangCMXCrossModalFusion2023}. TESR couples these ideas to explicit topographic context and reconstructs a single rectified visual feature rather than treating terrain as a second fine-resolution representation.

Let $\mathbf{X}=\mathbf{E}_1$ and $\mathbf{U}=\mathbf{T}_1$ be the aligned shallow visual and terrain features. A fixed channel-wise Haar transform decomposes the visual feature into one low-frequency and three directional high-frequency subbands \citep{mallat1989multiresolution}:

\begin{equation}
(\mathbf{X}_{LL},\mathbf{X}_{LH},\mathbf{X}_{HL},\mathbf{X}_{HH})
=\operatorname{DWT}_{\mathrm{Haar}}(\mathbf{X}).
\end{equation}

A stride-2 terrain projection produces $\mathbf{T}_f$. A two-layer predictor maps $\mathbf{T}_f$ to spatial FiLM parameters $\boldsymbol{\gamma}$ and $\boldsymbol{\beta}$, which rectify the low-frequency semantics \citep{perez2018film}:

\begin{equation}
\widetilde{\mathbf{X}}_{LL}
=\left(1+\tanh\boldsymbol{\gamma}\right)\odot\mathbf{X}_{LL}
+\boldsymbol{\beta}.
\end{equation}

The multiplicative factor lies in $(0,2)$. For the high-frequency path, the degree-valued slope is bilinearly resized to the wavelet grid as $\mathbf{S}_w$. High- and low-slope priors use fixed thresholds of $10^\circ$ and $5^\circ$, respectively, and their transition widths satisfy $\tau_h,\tau_l>0$. Low-slope enhancement additionally requires compatibility with the visual low-frequency feature:

\begin{equation}
\begin{array}{@{}l@{\;}c@{\;}l@{}}
\mathbf{P}_{h} &=& \sigma\!\left((\mathbf{S}_{w}-10^\circ)/\tau_h\right),\\
\mathbf{P}_{l} &=& \sigma\!\left((5^\circ-\mathbf{S}_{w})/\tau_l\right),\\
\mathbf{C}_{r} &=& \sigma\!\left(\phi_r(\mathbf{X}_{LL})\right).
\end{array}
\end{equation}

Let $a_h=\operatorname{softplus}(w_h)$ and $a_l=\operatorname{softplus}(w_l)$ be positive learnable strengths. The residual scale satisfies $b>0$. TESR combines their asymmetric physical energy with a bounded learned terrain residual. The factor $1-\mathbf{P}_h$ attenuates this residual on steep terrain so that it cannot override the high-slope anchor:

\begin{equation}
\begin{array}{@{}l@{\;}c@{\;}l@{}}
\mathbf{E}_{\mathrm{phys}} &=&
a_h\mathbf{P}_h-a_l\mathbf{P}_l\odot\mathbf{C}_r,\\
\mathbf{E}_{\mathrm{res}} &=&
b(1-\mathbf{P}_h)\odot\tanh\!\left(\phi_{\mathrm{hf}}(\mathbf{T}_f)\right),\\
\mathbf{E} &=& \mathbf{E}_{\mathrm{phys}}+\mathbf{E}_{\mathrm{res}}.
\end{array}
\end{equation}

Positive energy favors suppression and negative energy permits compatible enhancement. With temperature $T_E=0.35$, the energy is mapped monotonically to $\mathbf{G}=\sigma(-\mathbf{E}/T_E)$ and then smoothed by size-preserving $3\times3$ average pooling to obtain $\overline{\mathbf{G}}$. The modulation amplitude satisfies $\alpha>0$. Centering this gate around a neutral gain of one gives

\begin{equation}
\mathbf{A}_{\mathrm{hf}}=1+\alpha(2\overline{\mathbf{G}}-1).
\end{equation}

The same gain regulates all three directional subbands. For
$q\in\{LH,HL,HH\}$, this shared operation is

\begin{equation}
\widetilde{\mathbf{X}}_{q}
=\mathbf{A}_{\mathrm{hf}}\odot\mathbf{X}_{q}.
\end{equation}

Finally, the fixed inverse transform reconstructs the sole TESR output, which retains the spatial size and channel count of $\mathbf{X}$:

\begin{equation}
\mathbf{E}_1'=\operatorname{iDWT}_{\mathrm{Haar}}
(\widetilde{\mathbf{X}}_{LL},\widetilde{\mathbf{X}}_{LH},
\widetilde{\mathbf{X}}_{HL},\widetilde{\mathbf{X}}_{HH}).
\end{equation}

\subsection{Topography-guided Paddy Structure Decoder}
\label{subsec:tpsd}

\begin{figure*}[t]
\centering
\includegraphics[width=\textwidth]{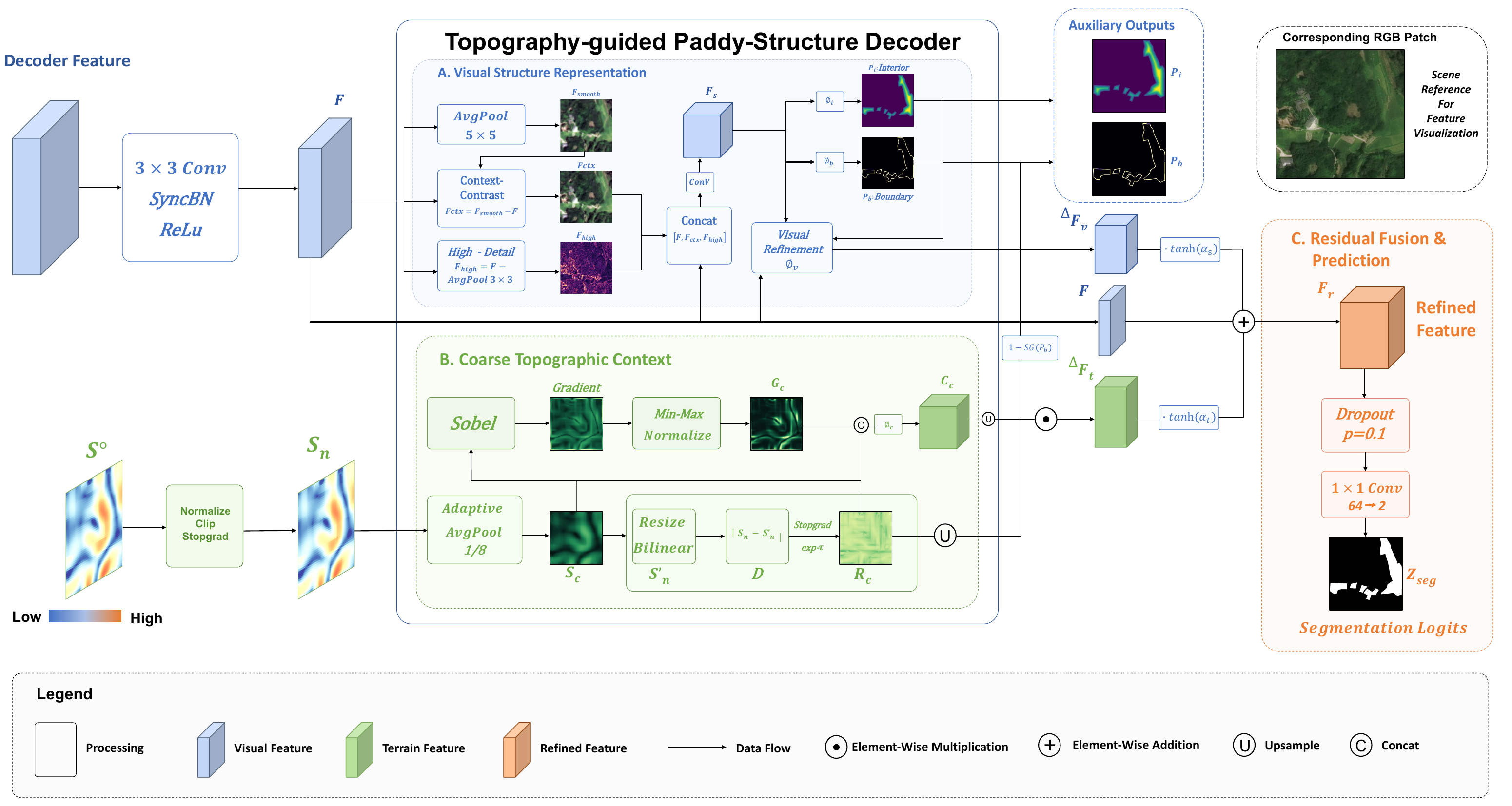}
\caption{Topography-guided Paddy Structure Decoder (TPSD). The full-resolution decoder feature is first transformed into the base feature $\mathbf{F}$. (A) The visual branch derives contextual and high-detail contrasts from $\mathbf{F}$ to predict interior-depth and boundary maps, which are fed back through visual refinement to form $\Delta\mathbf{F}_{\mathrm{v}}$. (B) The topographic branch combines coarse slope $\mathbf{S}_{c}$, normalized Sobel magnitude $\mathbf{G}_{c}$, and cross-scale consistency $\mathbf{R}_{c}$ to construct coarse context $\mathbf{C}_{c}$. The context and consistency paths are upsampled separately and combined with the boundary-free propagation mask to form $\Delta\mathbf{F}_{\mathrm{t}}$. (C) Learnable $\tanh$-scaled visual and topographic residuals are added to $\mathbf{F}$ before dropout and the final $1\times1$ two-class classifier. In the diagram, $P_i$ and $P_b$ denote the auxiliary interior-depth and boundary probabilities supervised during training.}
\label{fig:tpsd}
\end{figure*}

TPSD couples a decoder-derived visual structural residual with an explicit topographic residual so that they jointly refine rice--background boundaries and region interiors (Fig.~\ref{fig:tpsd}). Existing agricultural-field decoders improve geometry through joint mask, boundary, and distance-map prediction or boundary--semantic fusion \citep{long2022delineation,li2023using,shunying2023bsnet}. TPSD extends this structure-aware strategy by using coarse topographic context as an active refinement source, with boundary, interior, and cross-scale cues regulating how that context participates. Paddy structure here denotes semantic boundaries and erosion-derived interior depth, not cadastral or instance boundaries.

TPSD first transforms the full-resolution decoder output into a 64-channel base feature $\mathbf{F}$. Size-preserving average pooling provides a broad-context contrast $\mathbf{F}_{\mathrm{ctx}}=\operatorname{AvgPool}_{5\times5}(\mathbf{F})-\mathbf{F}$ and a fine-detail contrast $\mathbf{F}_{\mathrm{high}}=\mathbf{F}-\operatorname{AvgPool}_{3\times3}(\mathbf{F})$. These contrasts are fused into the shared structural feature

\begin{equation}
\mathbf{F}_{\mathrm{s}}=\phi_{\mathrm{s}}
([\mathbf{F},\mathbf{F}_{\mathrm{ctx}},\mathbf{F}_{\mathrm{high}}]).
\end{equation}

Independent heads predict boundary and interior-depth probabilities as $\mathbf{P}_{b}=\sigma(\phi_b(\mathbf{F}_{\mathrm{s}}))$ and $\mathbf{P}_{\mathrm{int}}=\sigma(\phi_{\mathrm{int}}(\mathbf{F}_{\mathrm{s}}))$, respectively. Both predictions are fed back with the shared feature to form the visual structural residual

\begin{equation}
\Delta\mathbf{F}_{\mathrm{v}}=\phi_{\mathrm{v}}
([\mathbf{F},\mathbf{F}_{\mathrm{s}},\mathbf{P}_{b},\mathbf{P}_{\mathrm{int}}]).
\end{equation}

The topographic path first defines $\mathbf{S}_{n}$ as the degree-valued slope clipped at $30^\circ$, divided by $30^\circ$ to obtain values in $[0,1]$, and detached from gradient flow. Adaptive average pooling of $\mathbf{S}_{n}$ to $H/8\times W/8$ then gives the coarse slope map $\mathbf{S}_{c}$. Bilinearly resizing $\mathbf{S}_{c}$ to $H\times W$ produces its full-resolution reconstruction $\widehat{\mathbf{S}}_{n}$. The element-wise absolute difference between $\mathbf{S}_{n}$ and $\widehat{\mathbf{S}}_{n}$ defines the reconstruction discrepancy $\boldsymbol{\Delta}_{s}$, and adaptive average pooling of $\boldsymbol{\Delta}_{s}$ to $H/8\times W/8$ yields its coarse-scale counterpart $\boldsymbol{\Delta}_{s,c}$. The deterministic cross-scale consistency weight is

\begin{equation}
\mathbf{R}_{c}=\operatorname{stopgrad}
\left(\exp(-\tau\boldsymbol{\Delta}_{s,c})\right),
\qquad \tau=6.
\end{equation}

This consistency measure differs from DEM accuracy: larger values indicate closer cross-scale slope reconstruction, not higher absolute elevation precision. Per-sample min--max-normalized Sobel magnitude on $\mathbf{S}_c$ gives the coarse transition map $\mathbf{G}_c$. These complementary topographic cues produce a learned coarse context:

\begin{equation}
\mathbf{C}_{c}=\phi_t
([\mathbf{S}_{c},\mathbf{G}_{c},\mathbf{R}_{c}]).
\end{equation}

The context and consistency maps are bilinearly resized to full resolution as $\mathbf{C}$ and $\mathbf{R}_{\mathrm{topo}}$. Boundary probability supplies a detached propagation mask, allowing the coarse context to refine region interiors while respecting predicted structure:

\begin{equation}
\Delta\mathbf{F}_{\mathrm{t}}
=\mathbf{C}\odot\mathbf{R}_{\mathrm{topo}}
\odot\left(1-\operatorname{stopgrad}(\mathbf{P}_{b})\right).
\end{equation}

Learnable $\tanh$-scaled coefficients inject both complementary residuals into the base feature:

\begin{equation}
\mathbf{F}_{\mathrm{r}}=\mathbf{F}
+\tanh(\alpha_{\mathrm{s}})\Delta\mathbf{F}_{\mathrm{v}}
+\tanh(\alpha_{\mathrm{t}})\Delta\mathbf{F}_{\mathrm{t}}.
\end{equation}

Dropout followed by a $1\times1$ classifier maps
$\mathbf{F}_{\mathrm{r}}$ to the final two-class logits
$\mathbf{Z}_{\mathrm{seg}}$.

\subsection{Training objectives}
\label{subsec:objectives}

Let $\mathbf{Y}\in\{0,1,255\}^{H\times W}$ encode background, rice, and ignored pixels. We denote the valid-pixel set by $\Omega_v$, the rice label at pixel $i\in\Omega_v$ by $y_i\in\{0,1\}$, the valid mask by $\mathbf{M}_{\mathrm{valid}}=[\mathbf{Y}\neq255]$, and the rice mask by $\mathbf{M}_{\mathrm{rice}}=[\mathbf{Y}=1]$. The overall training objective is

\begin{equation}
\label{eq:total_loss}
\mathcal{L}=\mathcal{L}_{\mathrm{TCE}}+\mathcal{L}_{\mathrm{Dice}}
+0.5\mathcal{L}_{\mathrm{boundary}}
+0.2\mathcal{L}_{\mathrm{interior}}.
\end{equation}

For degree-valued slope $s_i^\circ$ and logits $\mathbf{z}_i\in\mathbb{R}^{2}$, high- and low-slope memberships are defined by

\begin{equation}
\begin{array}{@{}l@{\;}c@{\;}l@{}}
h_i &=& \sigma\!\left(k(s_i^\circ-\theta_h)\right),\quad \theta_h=10^\circ,\\
l_i &=& \sigma\!\left(k(\theta_l-s_i^\circ)\right),\quad \theta_l=5^\circ.
\end{array}
\end{equation}

Both memberships vary smoothly around their respective thresholds rather than imposing hard partitions. The final configuration uses $k=0.75$ for both. The terrain-aware pixel weight is

\begin{equation}
w_i=1+w_{\mathrm{fp}}(1-y_i)h_i+w_{\mathrm{fn}}y_i l_i.
\end{equation}

Here, $w_{\mathrm{fp}}$ and $w_{\mathrm{fn}}$ independently control emphasis on high-slope background and low-slope rice. The final configuration sets $w_{\mathrm{fp}}=1.0$ and $w_{\mathrm{fn}}=0.25$. The Terrain-Guided Cross-Entropy Loss (TCE) is

\begin{equation}
\mathcal{L}_{\mathrm{TCE}}
=\frac{1}{|\Omega_v|}\sum_{i\in\Omega_v}
w_i\operatorname{CE}(\mathbf{z}_i,y_i).
\end{equation}

The terrain-dependent weights are computed independently of prediction confidence and emphasize steep-slope clutter and low-slope omissions. With rice probability $p_i=\operatorname{Softmax}(\mathbf{z}_i)_1$, the final configuration disables additional low-slope positive weighting in the Dice term and uses the standard foreground Dice loss \citep{milletari2016vnet}:

\begin{equation}
\mathcal{L}_{\mathrm{Dice}}=1-
\frac{2\sum_{i\in\Omega_v}p_i y_i+\epsilon}
{\sum_{i\in\Omega_v}p_i+\sum_{i\in\Omega_v}y_i+\epsilon}.
\end{equation}

Multi-task agricultural field delineation has used boundary supervision alongside semantic prediction \citep{long2022delineation}. Let $\mathbf{D}_r=\operatorname{Dilate}_{r=2}(\mathbf{M}_{\mathrm{rice}})$ and $\mathbf{E}_r=\operatorname{Erode}_{r=2}(\mathbf{M}_{\mathrm{rice}})$. Their difference generates the semantic boundary target directly from the rice mask:

\begin{equation}
\mathbf{B}^{\mathrm{gt}}
=(\mathbf{D}_r-\mathbf{E}_r)\odot\mathbf{M}_{\mathrm{valid}}.
\end{equation}

This finite-width boundary requires no instance annotation. Starting from $\mathbf{M}_{\mathrm{rice}}^{(0)}=\mathbf{M}_{\mathrm{rice}}$, we apply eight successive $3\times3$ erosions to obtain $\mathbf{M}_{\mathrm{rice}}^{(k)}$. Averaging these erosion-survival masks gives a truncated morphological depth proxy rather than a Euclidean distance or instance annotation:

\begin{equation}
\mathbf{I}^{\mathrm{gt}}=\frac{1}{8}\sum_{k=1}^{8}
\mathbf{M}_{\mathrm{rice}}^{(k)}.
\end{equation}

To address boundary imbalance, foreground and background pixels are weighted by their inverse frequencies over valid pixels. Counts and the normalization denominator are lower-bounded by one for degenerate samples. The complete boundary objective combines this normalized weighted binary cross-entropy with a boundary Dice term:

\begin{equation}
\mathcal{L}_{\mathrm{boundary}}
=0.5\mathcal{L}_{\mathrm{WBCE}}+\mathcal{L}_{\mathrm{bDice}}.
\end{equation}

For interior probability $P_{\mathrm{int},i}$, valid pixels receive weight $0.25+0.75y_i$, giving rice pixels four times the background weight. Let $q_i=M_{\mathrm{valid},i}(0.25+0.75y_i)$ and let $\ell_{\mathrm{SL1}}$ denote the standard Smooth L1 penalty. The normalized interior objective is

\begin{equation}
\mathcal{L}_{\mathrm{interior}}
 =\frac{\sum_i q_i\ell_{\mathrm{SL1}}(P_{\mathrm{int},i}-I_i^{\mathrm{gt}})}
{\max(\sum_i q_i,1)}.
\end{equation}

The boundary and interior losses supervise predictions that are also fed back into the final segmentation feature.

\subsection{Evaluation metrics}
\label{subsec:evaluation_metrics}

Paddy rice is the positive class, and ignored pixels are excluded from all confusion counts. We report rice precision, recall, and $F_1$ to characterize commission, omission, and their harmonic balance. Rice IoU is the primary metric because it measures foreground overlap under class imbalance, while mIoU averages the rice and background IoUs.

For the complementary boundary metric, let $\mathcal{B}_{p}$ and $\mathcal{B}_{g}$ be the one-pixel rice--background contours extracted from the predicted and reference masks. With Euclidean distance $d(\cdot,\mathcal{B})$ to the nearest contour pixel and tolerance $\delta=2$ pixels, boundary precision and recall are

\begin{equation}
\begin{array}{@{}l@{\;}c@{\;}l@{}}
P_b &=& \displaystyle
\frac{\sum_{\mathbf{x}\in\mathcal{B}_{p}}
\mathbb{I}[d(\mathbf{x},\mathcal{B}_{g})\leq\delta]}
{|\mathcal{B}_{p}|},\\[2pt]
R_b &=& \displaystyle
\frac{\sum_{\mathbf{x}\in\mathcal{B}_{g}}
\mathbb{I}[d(\mathbf{x},\mathcal{B}_{p})\leq\delta]}
{|\mathcal{B}_{g}|}.
\end{array}
\end{equation}

The corresponding Boundary F1 is

\begin{equation}
F_1^{\mathrm{boundary}}=\frac{2P_bR_b}{P_b+R_b}.
\end{equation}

For the terrain diagnostic, let $\Omega_{0,\geq15^\circ}=\{i\in\Omega_v:y_i=0,\ s_i^\circ\geq15^\circ\}$ be the valid high-slope background set. The high-slope false-positive rate is

\begin{equation}
\mathrm{FPR}_{\geq15^\circ}=
\frac{\sum_{i\in\Omega_{0,\geq15^\circ}}
\mathbb{I}[\widehat{y}_i=1]}
{|\Omega_{0,\geq15^\circ}|},
\end{equation}

where $\widehat{y}_i$ is the predicted class. Thus, this diagnostic measures the fraction of valid background pixels at slopes of at least $15^\circ$ that are misclassified as rice. Boundary F1 and high-slope background FPR quantify structural delineation and steep-terrain commission errors, respectively, without adding further model objectives.

\section{Experiments and Analysis}

\label{sec:experiments}

\subsection{Experimental settings}

\label{subsec:experimental_settings}

All models use the splits in Section~\ref{subsec:dataset_split}. Training and validation-based checkpoint selection use the 2023 Area A data; evaluation uses its internal test set, held-out Area B, and the matched August 2024 test sets. The August 2024 data are inference-only: model weights, checkpoints, normalization statistics, and decision rules remain fixed from 2023. RGB-only and multimodal models use three and five channels, respectively. TRNet uses the same five underlying input channels as the other multimodal models; the unstandardized degree-valued slope is additionally retained from the same slope raster for physically defined modulation and loss weighting. All models are trained from scratch under a common protocol and evaluated without post-processing. Dual-Encoder U-Net, TRNet-Base, and TRNet are each trained five times with independent random seeds. All metrics for these models are reported as mean $\pm$ standard deviation over the five independent runs.

Training uses $512\times512$ patches with random scaling, cropping, and horizontal flipping. Stochastic gradient descent runs for 30,000 iterations with batch size 2, initial learning rate 0.003, momentum 0.9, and weight decay 0.0005. A polynomial schedule follows 1,000 warm-up iterations. Whole-image inference uses an NVIDIA GeForce RTX 3090.

Decoder and objective ablations also report rice--background Boundary F1 with 2-pixel tolerance and high-slope background FPR over pixels with slope $\geq15^{\circ}$. Settings are fixed within each ablation group. Rice IoU uses the main semantic evaluator; the other metrics are complementary diagnostics from the same checkpoint.

\subsection{Comparison with baseline methods}

\label{subsec:baseline_comparison}

The CNN baselines are U-Net \citep{ronnebergerUNetConvolutionalNetworks2015}, DeepLabV3+ \citep{chen2018encoderdecoder}, and OCRNet \citep{yuan2020object}. The Transformer baselines are SegFormer \citep{xie2021segformer} and GloTS \citep{liu2023rethinking}. The state-space-based baselines are SegMAN \citep{fu2025segman}, which combines sliding local attention with dynamic state-space modeling, and Samba \citep{zhu2024samba}. We also evaluate the remote-sensing model FADNet \citep{liu2024fadnet}. All baseline methods are reproduced from their official code releases under the unified experimental protocol described above. RGB-only and early-concatenation RGB--DEM--slope variants isolate the effect of added terrain channels; RGB+T denotes the latter input configuration. The original Dual-Encoder U-Net serves as a prior-art comparison. TRNet-Base retains our redesigned asymmetric backbone and decoder but replaces TESR and TPSD with the standard interaction and semantic head defined in Section~\ref{subsec:overall_architecture}; it serves as the controlled architectural baseline for all module ablations.

\begin{table*}[t]

\centering

\scriptsize

\setlength{\tabcolsep}{4pt}

\caption{Area A internal-test comparison. RGB+T denotes early RGB--DEM--slope concatenation; Separate denotes modality-specific encoder streams before cross-modal interaction. All five metrics for Dual-Encoder U-Net, TRNet-Base, and TRNet are reported as mean $\pm$ standard deviation over five runs; other baselines are single-run percentages.}

\label{tab:area_a_comparison}

\begin{tabular*}{\textwidth}{@{\extracolsep{\fill}}lcrrrrr}

\toprule

Method & Input/fusion & Rice IoU & Rice F1 & Rice Precision & Rice Recall & mIoU \tabularnewline

\midrule

DeepLabV3+~\citep{chen2018encoderdecoder} & RGB+T & 62.35 & 76.81 & 81.52 & 72.62 & 76.86 \tabularnewline

DeepLabV3+~\citep{chen2018encoderdecoder} & RGB & 62.87 & 77.20 & 81.20 & 73.59 & 77.14 \tabularnewline

FADNet~\citep{liu2024fadnet} & RGB & 63.01 & 77.31 & 84.15 & 71.49 & 77.38 \tabularnewline

OCRNet~\citep{yuan2020object} & RGB & 63.65 & 77.79 & 83.44 & 72.86 & 77.72 \tabularnewline

SegMAN~\citep{fu2025segman} & RGB+T & 65.05 & 78.82 & 83.13 & 74.94 & 78.54 \tabularnewline

GloTS~\citep{liu2023rethinking} & RGB+T & 66.38 & 79.79 & 84.40 & 75.66 & 79.39 \tabularnewline

FADNet~\citep{liu2024fadnet} & RGB+T & 66.52 & 79.89 & 79.84 & 79.95 & 79.24 \tabularnewline

GloTS~\citep{liu2023rethinking} & RGB & 66.60 & 79.95 & 82.31 & 77.73 & 79.42 \tabularnewline

OCRNet~\citep{yuan2020object} & RGB+T & 67.05 & 80.28 & 77.45 & 83.32 & 79.40 \tabularnewline

SegFormer~\citep{xie2021segformer} & RGB+T & 67.65 & 80.70 & 80.79 & 80.62 & 79.96 \tabularnewline

SegMAN~\citep{fu2025segman} & RGB & 68.07 & 81.01 & 82.48 & 79.58 & 80.31 \tabularnewline

SegFormer~\citep{xie2021segformer} & RGB & 69.73 & 82.16 & 87.44 & 77.49 & 81.51 \tabularnewline

Samba~\citep{zhu2024samba} & RGB & 73.75 & 84.89 & 81.42 & 88.68 & 83.66 \tabularnewline

U-Net~\citep{ronnebergerUNetConvolutionalNetworks2015} & RGB+T & 74.33 & 85.27 & 85.96 & 84.59 & 84.21 \tabularnewline

Samba~\citep{zhu2024samba} & RGB+T & 75.29 & 85.90 & 84.93 & 86.89 & 84.75 \tabularnewline

U-Net~\citep{ronnebergerUNetConvolutionalNetworks2015} & RGB & 75.48 & 86.03 & 84.73 & 87.36 & 84.86 \tabularnewline

Dual-Encoder U-Net~\citep{luDualEncoderUNet2023} & Separate & 75.95 $\pm$ 0.28 & 86.33 $\pm$ 0.20 & 86.17 $\pm$ 0.25 & 86.48 $\pm$ 0.18 & 85.20 $\pm$ 0.16 \tabularnewline

TRNet-Base & Separate & 75.99 $\pm$ 0.11 & 86.36 $\pm$ 0.14 & 82.50 $\pm$ 0.22 & 90.60 $\pm$ 0.16 & 85.07 $\pm$ 0.12 \tabularnewline

TRNet & Separate & \textbf{85.10 $\pm$ 0.26} & \textbf{91.95 $\pm$ 0.18} & \textbf{89.59 $\pm$ 0.21} & \textbf{94.43 $\pm$ 0.15} & \textbf{90.85 $\pm$ 0.17} \tabularnewline

\bottomrule

\end{tabular*}

\end{table*}

\begin{figure*}[t]

\centering

\includegraphics[width=\textwidth]{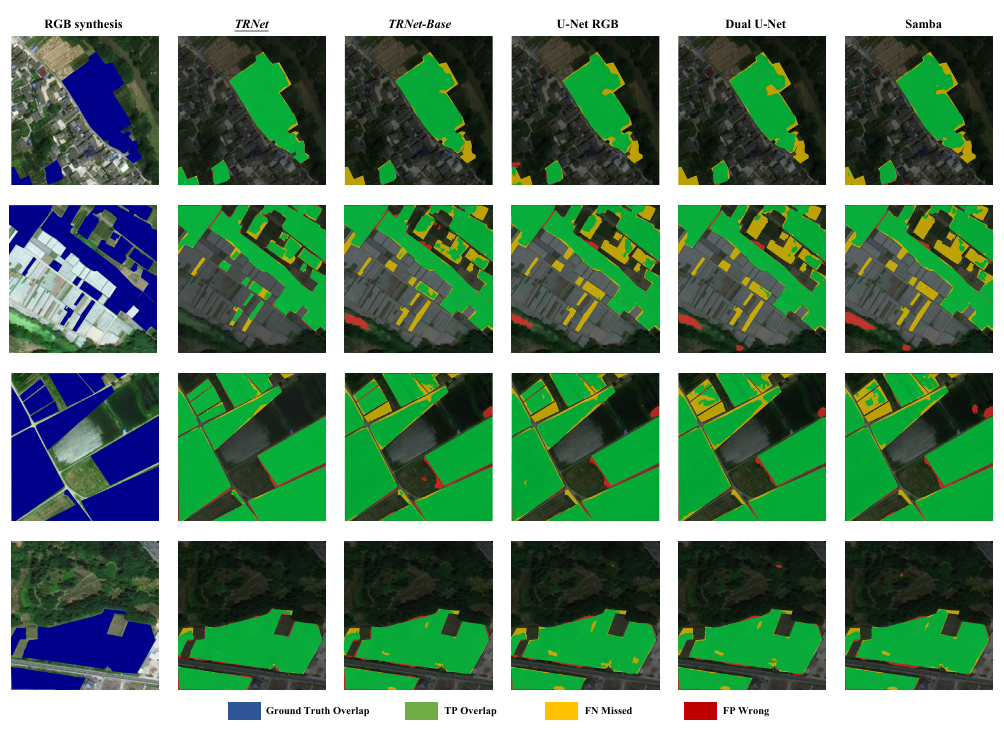}

\caption{Qualitative comparison on four representative Area A test scenes. Columns show the RGB image with ground-truth overlay, followed by error overlays for TRNet, TRNet-Base, RGB U-Net, the original Dual-Encoder U-Net, and Samba; colors denote true positives, false negatives, and false positives.}

\label{fig:baseline_qualitative}

\end{figure*}

On Area A, TRNet exceeds the original Dual-Encoder U-Net, TRNet-Base, and the strongest RGB-only baseline by 9.15, 9.11, and 9.62 Rice IoU points, respectively (Table~\ref{tab:area_a_comparison}). Relative to the original network, Rice precision and recall increase by 3.42 and 7.95 points, respectively. Relative to TRNet-Base, they increase by 7.09 and 3.83 points, indicating that TRNet improves both false-positive control and rice-region coverage. Figure~\ref{fig:baseline_qualitative} provides the corresponding qualitative comparison on representative Area A scenes.

Early terrain concatenation improves only three of eight architectures, yielding a negligible mean Rice IoU change of $+0.18$ points. TRNet's gain therefore reflects explicit terrain--visual interaction rather than additional channels alone.

\begin{table}[t]

\centering

\scriptsize

\setlength{\tabcolsep}{4pt}

\caption{Held-out Area B comparison. RGB+T denotes early concatenation; Separate denotes modality-specific encoder streams before cross-modal interaction. The table includes all multimodal models and the strongest Area A RGB-only model. All five metrics for Dual-Encoder U-Net, TRNet-Base, and TRNet are reported as mean $\pm$ standard deviation over five runs; other baselines are single-run percentages.}

\label{tab:area_b_comparison}

\resizebox{\columnwidth}{!}{%
\begin{tabular}{lcrrrrr}

\toprule

Method & Input/fusion & Rice IoU & Rice F1 & Rice Precision & Rice Recall & mIoU \tabularnewline

\midrule

DeepLabV3+~\citep{chen2018encoderdecoder} & RGB+T & 51.26 & 67.78 & 66.39 & 69.23 & 74.00 \tabularnewline

SegFormer~\citep{xie2021segformer} & RGB+T & 51.90 & 68.34 & 64.24 & 72.99 & 74.27 \tabularnewline

OCRNet~\citep{yuan2020object} & RGB+T & 53.05 & 69.33 & 63.94 & 75.70 & 74.86 \tabularnewline

GloTS~\citep{liu2023rethinking} & RGB+T & 53.24 & 69.49 & 61.66 & 79.58 & 74.88 \tabularnewline

SegMAN~\citep{fu2025segman} & RGB+T & 54.30 & 70.38 & 64.21 & 77.85 & 75.52 \tabularnewline

FADNet~\citep{liu2024fadnet} & RGB+T & 55.45 & 71.34 & 70.12 & 72.61 & 76.28 \tabularnewline

Samba~\citep{zhu2024samba} & RGB+T & 56.51 & 72.21 & 64.92 & 81.36 & 76.69 \tabularnewline

U-Net~\citep{ronnebergerUNetConvolutionalNetworks2015} & RGB+T & 57.08 & 72.68 & 62.70 & 86.44 & 76.92 \tabularnewline

U-Net~\citep{ronnebergerUNetConvolutionalNetworks2015} & RGB & 57.75 & 73.22 & 61.11 & 91.32 & 77.20 \tabularnewline

Dual-Encoder U-Net~\citep{luDualEncoderUNet2023} & Separate & 61.85 $\pm$ 0.19 & 76.43 $\pm$ 0.24 & 66.77 $\pm$ 0.35 & 89.34 $\pm$ 0.29 & 79.54 $\pm$ 0.22 \tabularnewline

TRNet-Base & Separate & 64.38 $\pm$ 0.14 & 78.33 $\pm$ 0.18 & 70.40 $\pm$ 0.28 & 88.28 $\pm$ 0.23 & 80.97 $\pm$ 0.19 \tabularnewline

TRNet & Separate & \textbf{80.68 $\pm$ 0.24} & \textbf{89.31 $\pm$ 0.21} & \textbf{86.06 $\pm$ 0.31} & \textbf{92.81 $\pm$ 0.26} & \textbf{89.78 $\pm$ 0.23} \tabularnewline

\bottomrule

\end{tabular}%
}

\end{table}

On Area B, TRNet exceeds the original Dual-Encoder U-Net and TRNet-Base by 18.83 and 16.30 Rice IoU points, respectively (Table~\ref{tab:area_b_comparison}). Relative to the original network, Rice precision and recall rise by 19.29 and 3.47 points. Relative to TRNet-Base, they rise by 15.66 and 4.53 points, showing that the complete model substantially improves false-positive control while further increasing rice coverage.

Area B differs in terrain, land cover, and rice prevalence but shares Area A's imagery source, preprocessing, and annotation protocol; it therefore measures cross-area transfer rather than independent external validation.

Overall, TRNet consistently outperforms the single-stream methods, the original Dual-Encoder U-Net, and TRNet-Base in Areas A and B. TRNet-Base exceeds the original network by 0.04 and 2.53 Rice IoU points in Areas A and B, respectively, which remains substantially smaller than the gains of the complete model. Redesigning the backbone alone therefore does not explain TRNet's improvement. Instead, TRNet improves false-positive control while maintaining or increasing rice coverage, with fewer fragmented rice strips and isolated false positives.

\subsection{Ablation studies}

\label{subsec:ablation}

\subsubsection{TESR encoder placement and frequency paths}

\begin{table}[t]

\centering

\scriptsize

\caption{Area A TESR ablation across encoder stages and frequency paths with TPSD enabled in all variants. LF-only retains low-frequency FiLM; HF-only retains high-frequency gating.}

\label{tab:tesr_ablation}

\resizebox{\columnwidth}{!}{%
\begin{tabular}{lrrrrr}

\toprule

Variant & Rice IoU & Rice F1 & Rice Precision & Rice Recall & mIoU \tabularnewline

\midrule

w/o & 76.79 & 86.87 & 79.82 & \textbf{95.30} & 85.42 \tabularnewline

E0 & 80.19 & 89.01 & 86.30 & 91.89 & 87.77 \tabularnewline

E1 & \textbf{85.10} & \textbf{91.95} & \textbf{89.59} & 94.43 & \textbf{90.85} \tabularnewline

E2 & 77.51 & 87.33 & 80.94 & 94.81 & 85.92 \tabularnewline

E3 & 77.42 & 87.27 & 80.58 & 95.18 & 85.84 \tabularnewline

E4 & 77.16 & 87.10 & 83.76 & 90.73 & 85.83 \tabularnewline

\midrule

\multicolumn{6}{l}{\emph{Frequency-path variants}} \tabularnewline

LF-only & 76.54 & 86.71 & 80.50 & 93.95 & 85.30 \tabularnewline

HF-only & 77.34 & 87.22 & 81.16 & 94.27 & 85.83 \tabularnewline

\bottomrule

\end{tabular}%
}

\end{table}

\begin{figure}[t]

\centering

\includegraphics[width=\columnwidth]{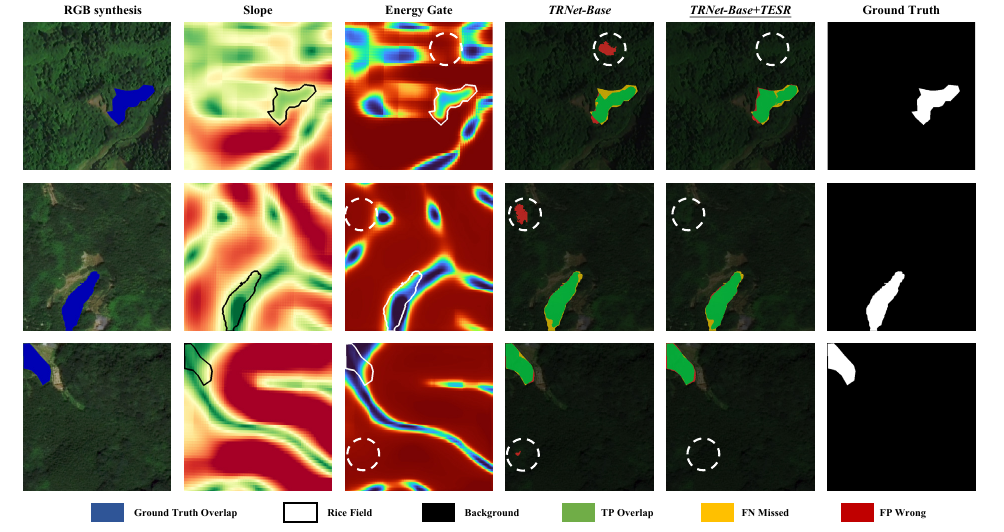}

\caption{TESR effects in three high-slope scenes. Columns show the RGB image with ground-truth overlay, slope, energy gate, error overlays for TRNet-Base and TRNet-Base + TESR, and ground truth. White circles link suppressed gate responses to removed baseline false positives.}

\label{fig:tesr_comparison}

\end{figure}

\begin{figure}[t]

\centering

\includegraphics[width=\columnwidth]{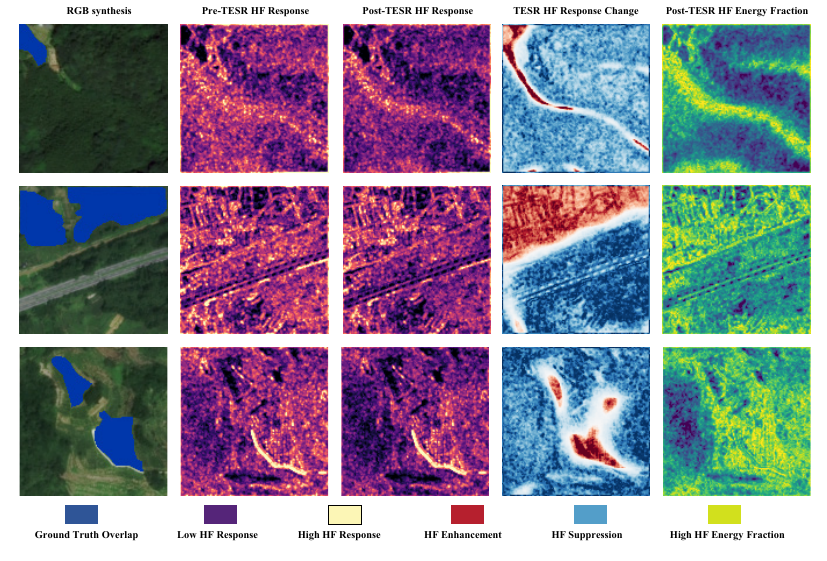}

\caption{High-frequency response analysis of TESR. From left to right, the columns show the RGB image with ground-truth overlay, the high-frequency response before TESR, the response after TESR, their difference (after minus before), and the high-frequency response ratio after TESR. Darker responses in the post-TESR panel indicate attenuation, whereas brighter regions in the difference and ratio panels indicate locally enhanced high-frequency responses. The high-slope non-rice regions are selectively darkened while compatible rice structure is retained or enhanced.}

\label{fig:tesr_hf_response}

\end{figure}

Table~\ref{tab:tesr_ablation} evaluates TESR placement across encoder stages $E0$--$E4$ and compares the complete module with its individual frequency paths on Area A. In the table, w/o denotes the variant without TESR, while E0--E4 denote the encoder stage at which TESR is inserted. TPSD remains enabled in every row, so the table isolates changes to TESR. LF-only and HF-only retain only the low- and high-frequency paths, respectively. At E1, TESR improves Rice IoU by 8.31 points over no TESR; precision rises from 79.82\% to 89.59\% while recall changes from 95.30\% to 94.43\%, indicating fewer false positives. E1 reaches 85.10\% Rice IoU and 90.85\% mIoU, beating the next-best E0 placement by 4.91 and 3.08 points, whereas E2--E4 yield 77.16--77.51\% Rice IoU. LF-only and HF-only reach 76.54\% and 77.34\% Rice IoU. Relative to them, E1 improves precision by 9.09 and 8.43 points and recall by 0.48 and 0.16 points, supporting complementary frequency paths. Figure~\ref{fig:tesr_comparison} compares TRNet-Base with its TESR-equipped variant in three high-slope, rice-sparse scenes. The white-circled regions show that TESR attenuates terrain-inconsistent high-slope responses that become isolated false positives in the variant without TESR, while sparse rice regions remain intact. The visualization therefore supports selective steep-terrain suppression rather than blanket removal. Figure~\ref{fig:tesr_hf_response} provides a direct frequency-domain view of TESR. Comparing the second and third columns shows that high-frequency responses in steep non-rice terrain become weaker after terrain-conditioned regulation. The difference map makes the spatial distribution of suppression and enhancement explicit, while the final ratio map highlights enhanced responses as brighter regions. Together, these views support TESR's asymmetric operation: it reduces terrain-induced high-frequency clutter and conditionally strengthens compatible rice-related detail rather than applying a uniform attenuation.

\subsubsection{TPSD and training objectives}

\begin{figure}[t]

\centering

\includegraphics[width=\columnwidth]{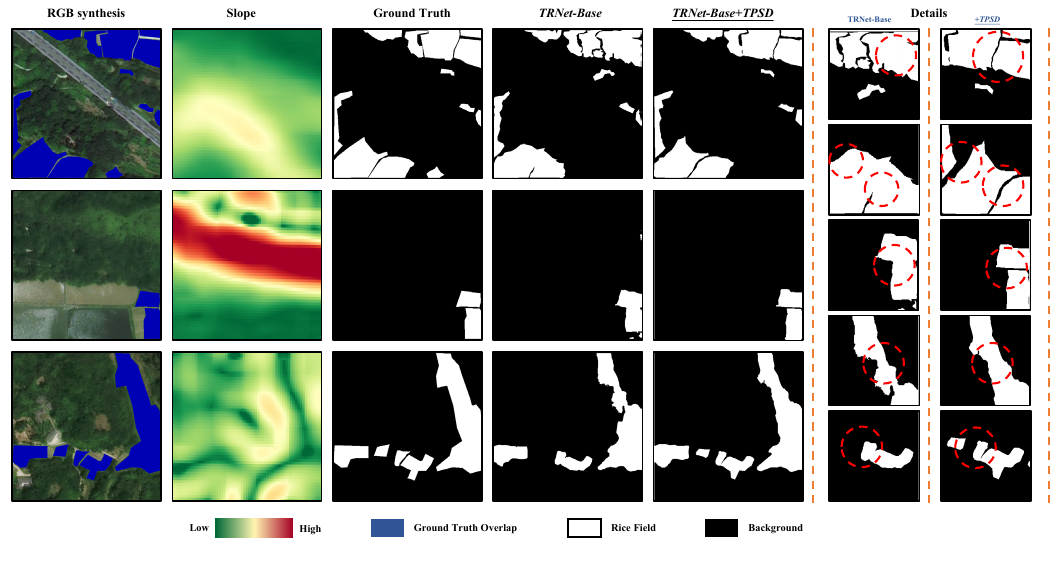}

\caption{Qualitative comparison of TRNet-Base and TRNet-Base + TPSD. Columns show the RGB image with ground-truth overlay, slope, ground truth, the two predictions, and paired detail crops. Red circles highlight boundary and interior differences; white and black indicate rice and background.}

\label{fig:tpsd_comparison}

\end{figure}

\begin{table}[t]

\centering

\scriptsize

\setlength{\tabcolsep}{4.8pt}

\caption{Area A TPSD and objective ablations. Boundary F1 uses 2-pixel tolerance; high-slope background FPR uses pixels at $\geq15^{\circ}$.}

\label{tab:tpsd_loss_ablation}

\resizebox{\columnwidth}{!}{%
\begin{tabular}{llrrr}

\toprule

Group & Variant & Rice IoU $\uparrow$ & Boundary F1 $\uparrow$ & High-slope background FPR $\downarrow$ \tabularnewline

\midrule

TPSD & TESR w/o TPSD & 81.48 & 49.90 & 3.11 \tabularnewline

& Boundary-only TPSD & 82.36 & 55.80 & 2.77 \tabularnewline

& Visual-structure TPSD & 84.55 & 60.15 & 2.31 \tabularnewline

& Full TPSD & \textbf{85.10} & \textbf{64.96} & \textbf{1.35} \tabularnewline

\midrule

Objective & CE + Dice & 76.58 & 50.72 & 2.95 \tabularnewline

& TCE + Dice & 78.23 & 51.67 & 1.83 \tabularnewline

& + Boundary supervision & 80.22 & 57.46 & 1.87 \tabularnewline

& + Interior supervision & 84.57 & 60.13 & 1.58 \tabularnewline

& Final balanced objective & \textbf{85.10} & \textbf{64.96} & \textbf{1.35} \tabularnewline

\bottomrule

\end{tabular}%
}

\end{table}

Table~\ref{tab:tpsd_loss_ablation} compares full and reduced TPSD variants with TESR and the final objective fixed. Objective variants retain the complete architecture and cumulatively replace CE with TCE, add boundary and interior supervision, and apply the balanced weights in Eq.~\eqref{eq:total_loss}. Full TPSD exceeds the TESR-equipped decoder by 3.62 Rice IoU and 15.06 Boundary F1 points while reducing high-slope background FPR from 3.11\% to 1.35\%. Against visual-structure TPSD, the stronger reduced variant, it adds 0.55 and 4.81 points and lowers FPR by 0.96 points. This supports complementary visual structure and topographic context. Figure~\ref{fig:tpsd_comparison} compares TRNet-Base with its TPSD-equipped variant. The red-circled crops show that TPSD restores more continuous rice--background contours and closes internal breaks, yielding regions closer to the reference. These comparisons concern semantic structure rather than instance or cadastral delineation. Replacing CE with TCE raises Rice IoU and Boundary F1 from 76.58\% and 50.72\% to 78.23\% and 51.67\%; Dice remains standard because additional low-slope positive weighting is disabled. Boundary and interior supervision raise them to 84.57\% and 60.13\%. High-slope background FPR declines overall despite an increase from 1.83\% to 1.87\% after boundary supervision. The final balanced objective reaches 85.10\% Rice IoU, 64.96\% Boundary F1, and 1.35\% FPR. Over the strongest intermediate variant, gains are 0.53 and 4.83 points, with 0.23-point lower FPR.

\subsubsection{Overall component ablation}

Table~\ref{tab:overall_ablation} summarizes the overall component ablation. Adding TESR to TRNet-Base improves Rice IoU by 5.49 points, whereas adding TPSD alone yields a 0.80-point gain. Enabling both modules improves TRNet-Base by 9.11 points and exceeds the single-module variants by 3.62 and 8.31 points, respectively, supporting their complementary use.

\begin{table}[t]
\centering
\scriptsize
\caption{Overall component ablation on Area A. A check mark denotes an enabled module; $\Delta$ is the Rice IoU change from TRNet-Base. Values are percentage points.}
\label{tab:overall_ablation}
\begin{tabular}{lccrr}
\toprule
Variant & TESR & TPSD & Rice IoU & $\Delta$ \tabularnewline
\midrule
TRNet-Base & -- & -- & 75.99 & -- \tabularnewline
TRNet-Base + TESR & $\checkmark$ & -- & 81.48 & +5.49 \tabularnewline
TRNet-Base + TPSD & -- & $\checkmark$ & 76.79 & +0.80 \tabularnewline
Full TRNet & $\checkmark$ & $\checkmark$ & \textbf{85.10} & \textbf{+9.11} \tabularnewline
\bottomrule
\end{tabular}
\end{table}

\subsection{Terrain-stratified and cross-area analysis}

\label{subsec:terrain_analysis}

\begin{figure*}[t]

\centering

\includegraphics[width=\textwidth]{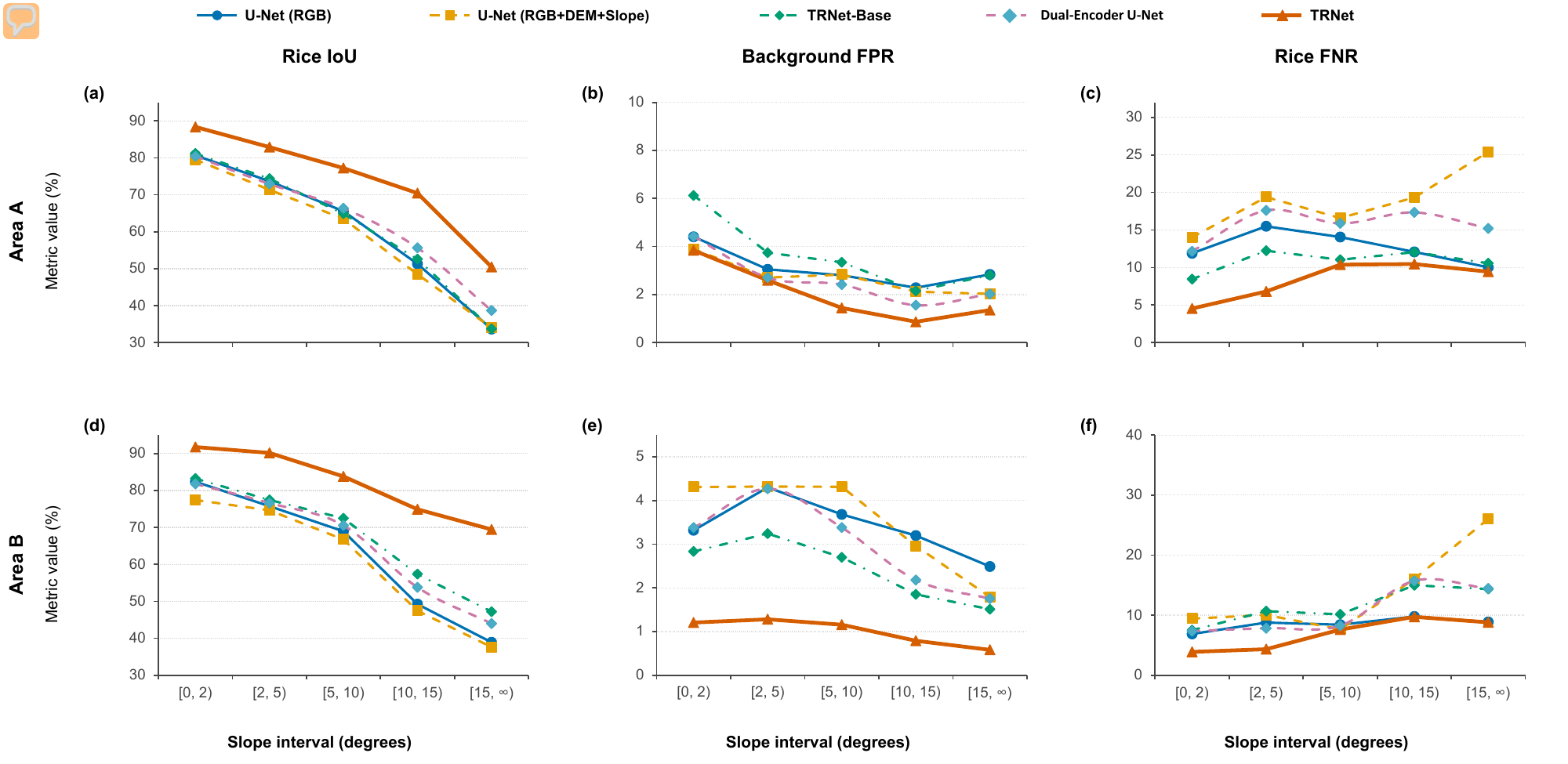}

\caption{Terrain-stratified performance in Area A (a--c) and Area B (d--f). Curves show Rice IoU, background FPR, and rice FNR for five models across five slope intervals using pooled valid pixels; label-255 pixels are excluded.}

\label{fig:slope_stratified}

\end{figure*}

We pool valid pixels from 147 Area A and 300 Area B test patches into $[0,2)$, $[2,5)$, $[5,10)$, $[10,15)$, and $[15,\infty)$ degree intervals. Aggregated confusion counts yield Rice IoU, background FPR, and rice FNR for RGB U-Net, RGB+T U-Net, TRNet-Base, the original Dual-Encoder U-Net, and TRNet (Fig.~\ref{fig:slope_stratified}).

High-slope background FPR evaluates whether the complete terrain-aware design exhibits the intended suppression behavior, while low-slope Rice IoU and FNR evaluate whether it reduces rice omissions.

In the lowest-slope interval, TRNet has the lowest FNR in both areas, indicating fewer low-slope rice omissions under the complete terrain-aware design.

TRNet has the highest Rice IoU in every interval. Its margin over the strongest comparator grows from 7.23 to 11.76 points from the lowest to highest slopes in Area A and from 8.54 to 22.25 points in Area B. At $\geq15^{\circ}$, its background FPR is 1.35\% and 0.58\%, below the baseline ranges of 2.02--2.84\% in Area A and 1.51--2.49\% in Area B. Rice FNR is 9.45\% and 8.80\%, also below all baselines in the two areas. The joint reductions in false positives and false negatives support improved steep-terrain discrimination rather than a trade-off between suppression and rice coverage.

The same ranking holds at $10^{\circ}$ and $15^{\circ}$ slope cutoffs. TRNet FPR is 1.15\% and 1.35\% in Area A and 0.64\% and 0.58\% in Area B, below the best baseline ranges of 1.83--2.02\% and 1.51--1.61\%, respectively. Rice FNR remains within 9.45--10.03\% and 8.80--9.22\%, and is lower than the corresponding baselines at both cutoffs. The joint reduction is therefore not tied to a single slope threshold.

Together, the lower high-slope background FPR and low-slope FNR show that the complete TRNet design reduces both steep-terrain false positives and low-slope rice omissions in both areas.

\subsection{Cross-year seasonal generalization}

\label{subsec:cross_year_seasonal}

We directly apply the models trained and selected on the 2023 data to the matched August 2024 images without fine-tuning. Area A compares the same test footprints across years and therefore primarily measures temporal appearance and planting changes. Area B retains the geographic holdout under the same cross-year seasonal shift, producing a joint spatial--temporal shift. RGB U-Net, the original Dual-Encoder U-Net, TRNet-Base, and TRNet are included to distinguish the effect of terrain-aware regulation from general multimodal fusion. In addition to Rice IoU, we pool pixels by the fixed reference slope and report high-slope background FPR at $\geq15^{\circ}$ and low-slope rice FNR below $2^{\circ}$.

\begin{table}[t]
\centering
\scriptsize
\setlength{\tabcolsep}{3.2pt}
\caption{Cross-year seasonal evaluation without adaptation using the July 2023 and August 2024 acquisitions. The upper panel reports Rice IoU, with the cross-year change $\Delta$ in parentheses; the lower panel reports August 2024 terrain-stratified diagnostics. A/B denote Areas A and B. Values for Dual-Encoder U-Net, TRNet-Base, and TRNet are mean $\pm$ standard deviation over five checkpoints; U-Net is single-run.}
\label{tab:cross_year_seasonal}
\resizebox{\columnwidth}{!}{%
\begin{tabular}{lrrrr}
\toprule
Method & A 2023 & A 2024 ($\Delta$) & B 2023 & B 2024 ($\Delta$) \tabularnewline
\midrule
U-Net (RGB) & 75.48 & $69.62\;(-5.86)$ & 57.75 & $50.44\;(-7.31)$ \tabularnewline
Dual-Encoder U-Net & $75.95\pm0.28$ & $(71.34\pm0.34;-4.61)$ & $61.85\pm0.19$ & $(56.72\pm0.31;-5.13)$ \tabularnewline
TRNet-Base & $75.99\pm0.11$ & $(71.86\pm0.29;-4.13)$ & $64.38\pm0.14$ & $(59.62\pm0.28;-4.76)$ \tabularnewline
TRNet & $85.10\pm0.26$ & $\mathbf{(82.04\pm0.31;-3.06)}$ & $80.68\pm0.24$ & $\mathbf{(76.12\pm0.35;-4.56)}$ \tabularnewline
\midrule
\multicolumn{5}{l}{\emph{2024 terrain-stratified diagnostics (\%)}} \tabularnewline
Method & FPR A $\downarrow$ & FPR B $\downarrow$ & FNR A $\downarrow$ & FNR B $\downarrow$ \tabularnewline
\midrule
U-Net (RGB) & $3.28$ & $2.71$ & $14.82$ & $15.63$ \tabularnewline
Dual-Encoder U-Net & $2.94\pm0.12$ & $2.26\pm0.15$ & $13.65\pm0.31$ & $13.92\pm0.38$ \tabularnewline
TRNet-Base & $2.62\pm0.10$ & $2.08\pm0.13$ & $12.91\pm0.27$ & $12.74\pm0.32$ \tabularnewline
TRNet & $\mathbf{1.58\pm0.08}$ & $\mathbf{0.82\pm0.07}$ & $\mathbf{8.97\pm0.22}$ & $\mathbf{9.41\pm0.28}$ \tabularnewline
\bottomrule
\end{tabular}%
}
\end{table}

All methods decline on the August 2024 images, confirming that the matched acquisition introduces a nontrivial cross-year seasonal shift (Table~\ref{tab:cross_year_seasonal}). TRNet decreases by 3.06 Rice IoU points in Area A and 4.56 points in Area B, compared with reductions of 4.13--5.86 and 4.76--7.31 points for the three comparators. Despite the larger joint shift in Area B, TRNet retains 82.04\% and 76.12\% Rice IoU and exceeds TRNet-Base by 10.18 and 16.50 points in Areas A and B, respectively. The complete model therefore maintains its advantage rather than relying on appearance conditions specific to the 2023 acquisition.

The terrain-stratified diagnostics show that this persistence is not obtained through uniformly conservative prediction. Relative to TRNet-Base, TRNet reduces 2024 high-slope background FPR by 1.04 points in Area A and 1.26 points in Area B, while reducing low-slope rice FNR by 3.94 and 3.33 points. TRNet thus reduces both steep-terrain false positives and low-slope rice omissions under cross-year seasonal change. The somewhat larger Rice IoU decline in Area B is consistent with its simultaneous geographic, prevalence, and seasonal shifts, so the result should be interpreted as joint spatial--temporal transfer rather than a pure seasonal effect.

\subsection{Sensitivity to DEM resolution and elevation noise}

\label{subsec:dem_sensitivity}

We evaluate input sensitivity without retraining or fine-tuning the locked TRNet configuration. Let $z_5$ denote the native 5-m DEM. For the resolution test, $z_5$ is area-averaged to 10 or 30 m and bilinearly resampled to the original grid before applying the same normalization used during training. Slope is recomputed from each degraded DEM rather than resampling the original slope map. For the noise test, we use $\tilde{z}_{\sigma,k}=z_5+\eta_{\sigma,k}$, where $\eta_{\sigma,k}$ is a zero-mean Gaussian elevation-error field with standard deviation $\sigma\in\{1,3,5\}$ m and a 10-m spatial correlation length. We evaluate the five independently trained checkpoints used for the main comparison; for each checkpoint and noise level, metrics are averaged over five independently generated error fields. RGB images, model weights, normalization statistics, and evaluation masks remain fixed, so changes relative to the clean 5-m input isolate sensitivity to topographic quality.

\begin{table}[t]
\centering
\scriptsize
\setlength{\tabcolsep}{3.5pt}
\caption{Sensitivity of the locked TRNet configuration to DEM resolution and elevation noise. Values are reported as mean $\pm$ standard deviation across five independently trained checkpoints. For each noisy condition, the metric for each checkpoint is first averaged over five independently generated elevation-error fields. Rice IoU changes from the clean 5-m DEM are given by $\Delta$ in parentheses. FPR A/B denotes high-slope background FPR in Areas A and B using the fixed reference-slope stratum.}
\label{tab:dem_sensitivity}
\resizebox{\columnwidth}{!}{%
\begin{tabular}{llccc}
\toprule
Test & DEM condition & Area A IoU $\uparrow$ & Area B IoU $\uparrow$ & FPR A/B $\downarrow$ \tabularnewline
\midrule
Reference & 5 m, clean & $85.10\pm0.26$ (--) & $80.68\pm0.24$ (--) & $1.35\pm0.06\,/\,0.58\pm0.05$ \tabularnewline
\midrule
Resolution & 10 m & $(84.72\pm0.28;-0.38)$ & $(80.21\pm0.27;-0.47)$ & $(1.48\pm0.07\,/\,0.66\pm0.06)$ \tabularnewline
& 30 m & $(83.61\pm0.33;-1.49)$ & $(78.96\pm0.36;-1.72)$ & $(1.89\pm0.10\,/\,0.91\pm0.08)$ \tabularnewline
\midrule
Elevation noise & $\sigma=1$ m & $(84.93\pm0.29;-0.17)$ & $(80.49\pm0.28;-0.19)$ & $(1.43\pm0.06\,/\,0.64\pm0.05)$ \tabularnewline
& $\sigma=3$ m & $(84.28\pm0.34;-0.82)$ & $(79.77\pm0.35;-0.91)$ & $(1.68\pm0.08\,/\,0.79\pm0.07)$ \tabularnewline
& $\sigma=5$ m & $(83.35\pm0.41;-1.75)$ & $(78.71\pm0.43;-1.97)$ & $(2.03\pm0.11\,/\,1.05\pm0.09)$ \tabularnewline
\bottomrule
\end{tabular}%
}
\end{table}

Table~\ref{tab:dem_sensitivity} indicates gradual rather than abrupt degradation as topographic quality decreases. Reducing DEM resolution from 5 to 10 m changes Rice IoU by only 0.38 and 0.47 points in Areas A and B, respectively. Even at 30 m, the reductions remain 1.49 and 1.72 points. The noise test shows the same monotonic pattern: $\sigma=1$ m has a negligible effect, while $\sigma=5$ m lowers Rice IoU by 1.75 points in Area A and 1.97 points in Area B. High-slope background FPR increases smoothly with both coarsening and noise, rather than exhibiting a failure threshold. At 30 m it remains below the best baseline FPR values in both areas (2.02\% and 1.51\%). Under $\sigma=5$ m noise, Area A reaches 2.03\%, approximately the best baseline value, whereas Area B remains lower at 1.05\%. These results support robustness to moderate DEM degradation while also showing that cleaner and finer topography provides consistently better terrain regulation.

The limited loss at 30 m is consistent with the role assigned to topography in TRNet. DEM supplies coarse topographic contextual evidence, whereas decoder-derived visual structural residuals retain the fine rice--background geometry. The similar trends in the internal and held-out areas further suggest that the response is not specific to one test split. Because these tests synthetically alter one DEM product, however, they do not substitute for evaluation with independently acquired elevation products that may contain source-specific bias, void filling, or georegistration errors.

\subsection{Integrated analysis}

\label{subsec:integrated_analysis}

In the evaluated setting, the results support using topography to regulate visual features rather than simply treating DEM and slope as additional input channels. Early concatenation of DEM and slope changed the mean Rice IoU of eight architectures by only $+0.18$ points, whereas TRNet produced consistent gains over the RGB-only baselines, the original Dual-Encoder U-Net, and TRNet-Base. The cross-year seasonal comparison preserves this ranking without adaptation, with TRNet retaining a 10.18-point advantage over TRNet-Base in Area A and a 16.50-point advantage under the joint shift in Area B. Within the TESR ablation, the advantage was concentrated at $E_1$; applying the same operation at deeper stages yielded substantially smaller gains. This pattern suggests that terrain-conditioned frequency rectification is most useful while the visual representation still retains field-scale spatial detail. The interpretation remains architectural rather than causal, but it agrees with the intended division of roles: RGB supplies the fine geometry, while terrain modifies responses that are less compatible with the local topographic context.

The structure and terrain diagnostics further clarify the source of the improvement. Relative to the TESR-equipped decoder, full TPSD increased Boundary F1 by 15.06 points and reduced high-slope background FPR from 3.11\% to 1.35\%. The complete model also retained low rice omission rates in the steepest interval while suppressing background responses in both test areas. These results do not imply that the upsampled $5$-m DEM delineates $0.5$-m field boundaries. Rather, they support a complementary mechanism in which predicted visual structure guides coarse topographic context to refine region interiors while preserving fine boundaries. This distinction is important in fragmented mountain agriculture, where elevation data provide terrain-scale compatibility cues but cannot replace the spatial detail of VHR imagery.

Area B provides a demanding held-out cross-area test, and the 2024 acquisition adds a controlled cross-year seasonal shift, but neither constitutes independent regional or cross-sensor validation. Both years share the sensor and preprocessing pipeline, while Area B changes terrain, land cover, and rice prevalence together. The current evidence is also limited to one county, two matched growing-season acquisitions, one DEM product, and binary semantic masks. Consequently, the results do not establish multi-year stability beyond the two evaluated years, cross-sensor transfer, cadastral parcel delineation, or robustness to source-specific DEM bias and georegistration errors. Future work should prioritize geographically independent regions, additional years, image sensors, and independently acquired elevation products, and quantify computational cost. Instance-aware assessment would additionally require field-level annotations that are not available in the present dataset.

\section{Conclusion}
\label{sec:conclusion}

This work presented TRNet, a topography guided framework for VHR paddy rice segmentation in mountainous and hilly regions. TRNet treats coarse topography as contextual guidance rather than fine spatial evidence, and combines Topographic Energy Spectral Rectification with a Topography Guided Paddy Structure Decoder to regulate terrain induced visual responses and refine fragmented rice structures. TRNet achieves Rice IoU scores of 85.10\% on the Area A internal test set and 80.68\% on the geographically held out Area B, exceeding the original Dual Encoder U Net by 9.15 and 18.83 percentage points, respectively. Without adaptation, it further retains Rice IoU scores of 82.04\% and 76.12\% on matched August 2024 imagery. Ablation, terrain stratified, boundary, seasonal, and DEM sensitivity analyses consistently show that TRNet reduces steep terrain false positives and low slope rice omissions while improving structural delineation. These results demonstrate that coarse topography can serve as a stable contextual prior for robust VHR paddy rice mapping and that explicitly addressing the spatial scale mismatch between optical and terrain modalities is more effective than direct feature fusion. Future work will extend the evaluation to broader regions, additional years and sensors, and independently acquired elevation products.

\bibliographystyle{IEEEtranN}
\bibliography{rice}

\end{document}